\PassOptionsToPackage{table}{xcolor} 
\documentclass[letterpaper]{article} 
\usepackage[preprint]{aaai2027}      
\usepackage[hyphens]{url}            
\usepackage{graphicx}                
\usepackage{natbib}                  
\usepackage{caption}                 
\usepackage{booktabs}             
\usepackage{multirow}             
\usepackage{amsmath}              
\usepackage{amssymb}              
\usepackage{pifont}               
\usepackage{makecell}
\usepackage{siunitx}
\usepackage{graphicx}
\usepackage{graphicx}
\usepackage{subcaption}

\graphicspath{{figures/}}

\title{Understanding Dynamic Scenes at Gigapixel Scale: Wide-Area Spatio-Temporal Perception from UAVs}

\author{
  Yuhang Zhu\equalcontrib,
  Meiyi Zhu\equalcontrib,
  Yunkai Dang,
  Zhangnan Li,
  Yuxuan Wang,
  Wenbin Li\corresponding,
  Hongbing Pan
}
\affiliations{
  State Key Laboratory for Novel Software Technology, Nanjing University
}

\begin{document}

\maketitle

\begin{abstract}
  UAV-borne imaging has advanced from megapixel to gigapixel sensors, shifting aerial perception from recognizing individual targets to understanding entire dynamic scenes.
We characterize this demand as \emph{Wide-area Spatio-temporal Scene Understanding (WSTU)}, which requires wide-area coverage, per-target resolution, and temporal continuity at once, a combination existing datasets lack.
To fill this gap, we introduce an ultra-\textbf{H}igh-resolution ($12768{\times}9564$) \textbf{A}irborne \textbf{R}emote-sensing \textbf{D}ataset(HARD) annotated at three levels for object detection, multi-object tracking, and scene-level visual question answering.
Ultra-high-resolution imagery raises per-frame processing time to seconds.
At that scale latency can no longer be ignored in evaluation.
Thus, we propose a latency-aware metric for multi-object tracking called streaming-HOTA (s-HOTA).
Extensive baseline experiments show how ultra-high-resolution processing reshapes each task.
For detection, the end-to-end pipeline affects accuracy and speed as much as the detector itself does.
For tracking, high latency charges the association axis far more unevenly than the detection axis, and association is where pipelines diverge.
As a result, the pipeline that performs best offline can lose its lead under s-HOTA.
For VQA, vision-language models remain weak at cross-frame identity binding and cannot transfer their single-frame gains to it.
Together these findings show that the baselines we evaluate fall short of WSTU.
HARD provides the data and the systematic baselines to advance it.
Our dataset is available at \url{https://huggingface.co/RL-MIND}.

\end{abstract}


\section{Introduction}

Remote sensing~\cite{satmtb2023,ding2021dota,m3ot,uavobb} has long been the primary means of observing ground scenes at scale, with satellite and aerial imagery supporting applications from urban management and traffic monitoring to disaster response.
Among remote-sensing platforms, UAVs stand out for flexible low-altitude on-demand observation. 
Recent advances \cite{panda2020} in imaging technology have raised sensor resolution from the megapixel to the gigapixel class, so a single aerial frame can now cover a wide region while preserving per-object detail.
This shift has driven UAV vision from early object detection toward higher-level scene understanding.
The demand is no longer to localize isolated targets but to describe how the whole scene evolves.
We characterize this emerging demand as Wide-area Spatio-temporal Scene Understanding (WSTU).
WSTU requires a perception system to reason over wide regions spanning hundreds of meters and to track the objects within them continuously over time. 
Beyond this, the system should interpret the state of the scene as a whole rather than only its individual objects.
We therefore characterize WSTU by two properties that the data should provide.
A single frame should be large enough to cover the entire scene. 
And consecutive frames should be close enough in time to reveal how those targets move.

\begin{figure}[t]
  \centering
  \includegraphics[width=0.9\columnwidth]{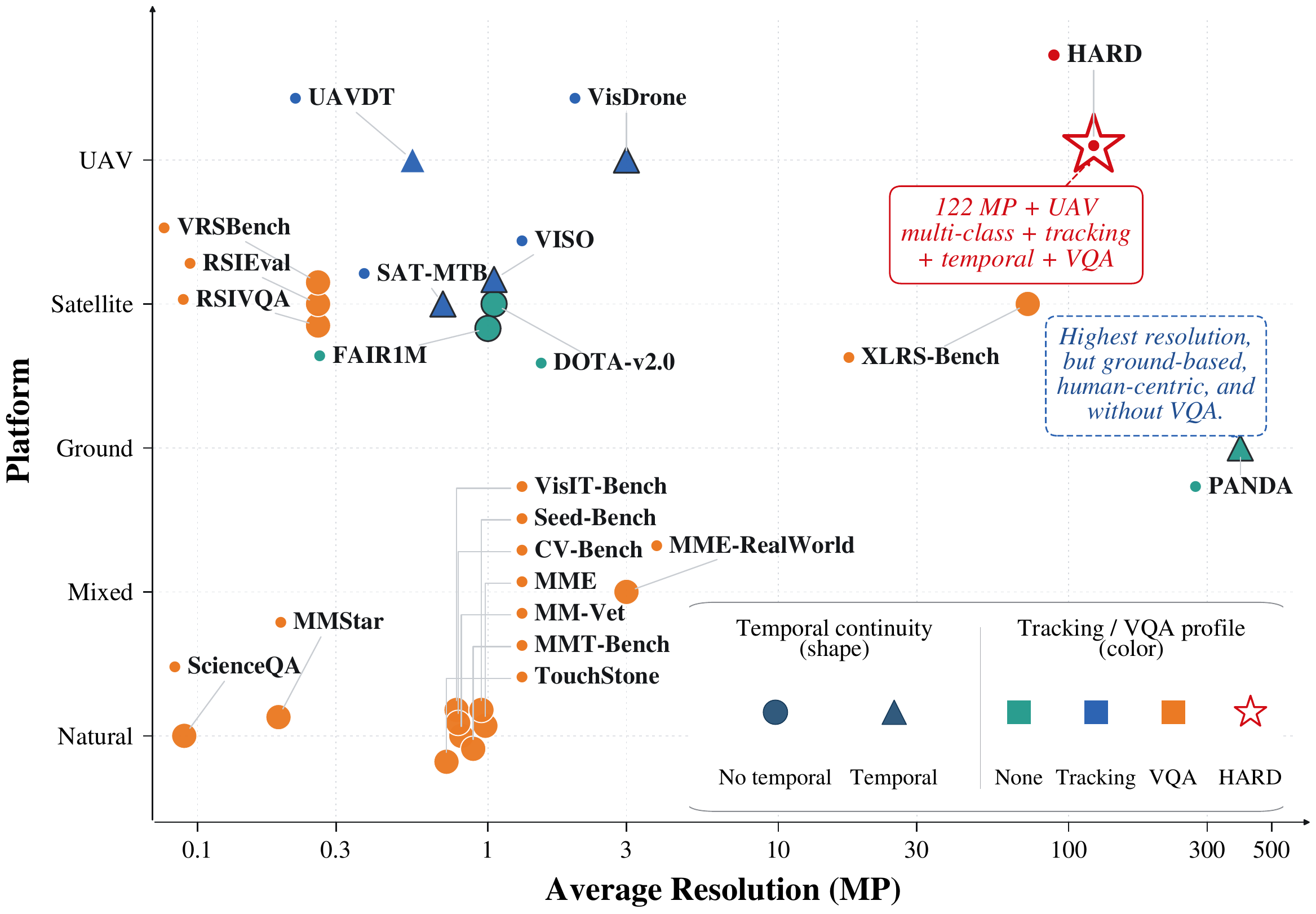}
  \caption{The dataset landscape. Average frame resolution is plotted against acquisition platform. Marker shape indicates temporal continuity and colour indicates whether tracking or VQA annotation is provided. HARD is the only benchmark that combines a UAV platform, 122\,MP frames, temporal continuity, multi-class tracking and scene-level VQA.}
  \label{fig:dataset_comparison}
\end{figure}

Figure~\ref{fig:dataset_comparison} compares representative existing datasets on the properties that WSTU requires.
UAV benchmarks~\cite{du2018unmanned,zhu2021detection} offer wide views but too little resolution to keep targets identifiable.
Gigapixel datasets~\cite{panda2020} reach high resolution only from fixed ground vantage points and lack a UAV's downward geometry.
Satellite datasets~\cite{viso2021,satmtb2023,ding2021dota,sun2022fair1m} provide spatial coverage but lack the resolution and temporal continuity that dynamic scene understanding needs.
More fundamentally, these benchmarks are built around individual objects.
They ask where each box is and how each trajectory continues but say almost nothing about the region as a whole.
Even recent remote-sensing VQA benchmarks~\cite{li2024vrsbench,hu2023rsgpt,rsivqa,wang2025xlrsbench} work on single still images rather than on wide areas that change over time.
For ultra-high-resolution images, the tiling and fusion pipeline around a detector pushes per-frame latency from milliseconds into seconds. 
Standard tracking metrics score every prediction against its own frame.
The staleness that latency induces is therefore never charged.
They reward offline accuracy and ignore whether a tracker can keep pace with the stream.

To close these gaps, we introduce the ultra-\textbf{H}igh-resolution \textbf{A}irborne \textbf{R}emote-sensing \textbf{D}ataset (HARD).
It is a UAV benchmark that pairs gigapixel-class frames with continuous temporal annotation for wide-area spatio-temporal scene understanding.
HARD comprises 3{,}549 frames at a unified 122\,MP resolution ($12768{\times}9564$), flight altitudes from 50 to 355\,m, and four flight modes.
HARD provides annotations for three tasks that progress from static individual objects to the dynamic scene.
Object detection (Level~1) builds on 138{,}830 expert-audited bounding boxes over four categories.
Multi-object tracking (Level~2) extends every box with a cross-frame instance identity.
Scene-level VQA (Level~3) rests on 1{,}565 manually verified question--answer pairs spanning eight question types.
In addition to the dataset, we design a latency-aware tracking metric called s-HOTA that evaluates each tracking pipeline online. 
This fills a gap left by offline metrics when assessing methods on ultra-high-resolution image sequences.
Based on this benchmark and metric, we conduct extensive experiments across all three levels, establishing a suite of baselines and laying a foundation for future WSTU research.

Extensive experiments across the three levels lead to four findings.
(F1)~Wide-area detection at full resolution is shaped as much by the tiling pipeline as by the detector.
(F2)~Streaming latency charges the association axis far more unevenly than the detection axis.
(F3)~Which tracking pipeline performs best under streaming is governed by the load ratio $\rho=\Delta t/P$.
A pipeline that ranks first under offline evaluation can see its lead weaken or even disappear once it is measured by s-HOTA.
(F4)~Zero-shot VLM failures separate into two axes, single-frame perception and cross-frame identity binding, and progress on one does not transfer to the other.
Models spread over a 37-point range on the first, while eleven of twelve stay within a 13-point band near chance on the second.
Our main contributions are as follows:

\begin{itemize}
  \item We construct a temporal UAV benchmark (HARD), which is the first to jointly provide wide-area coverage, ultra-high resolution, and temporal continuity. It contains 3{,}549 frames at a unified 122\,MP resolution, 138{,}830 boxes with cross-frame identities, and 1{,}565 scene-level QA pairs.
  \item We propose streaming-HOTA, a latency-aware extension of HOTA that charges each tracking pipeline for its end-to-end latency, remedying the zero-latency assumption of offline tracking metrics.
  \item We evaluate twenty detectors, four trackers, and twelve vision-language models across the three levels. These experiments establish a suite of baselines on HARD and provide reference points for future WSTU research.
  \item Our analysis traces how ultra-high-resolution processing shifts the bottleneck of each task. It identifies the factors that govern pipeline performance and turns them into concrete targets for future work on HARD.
\end{itemize}


\section{Related Work}
\label{sec:related}

\begin{figure*}[t]
  \centering
  \includegraphics[width=0.88\textwidth]{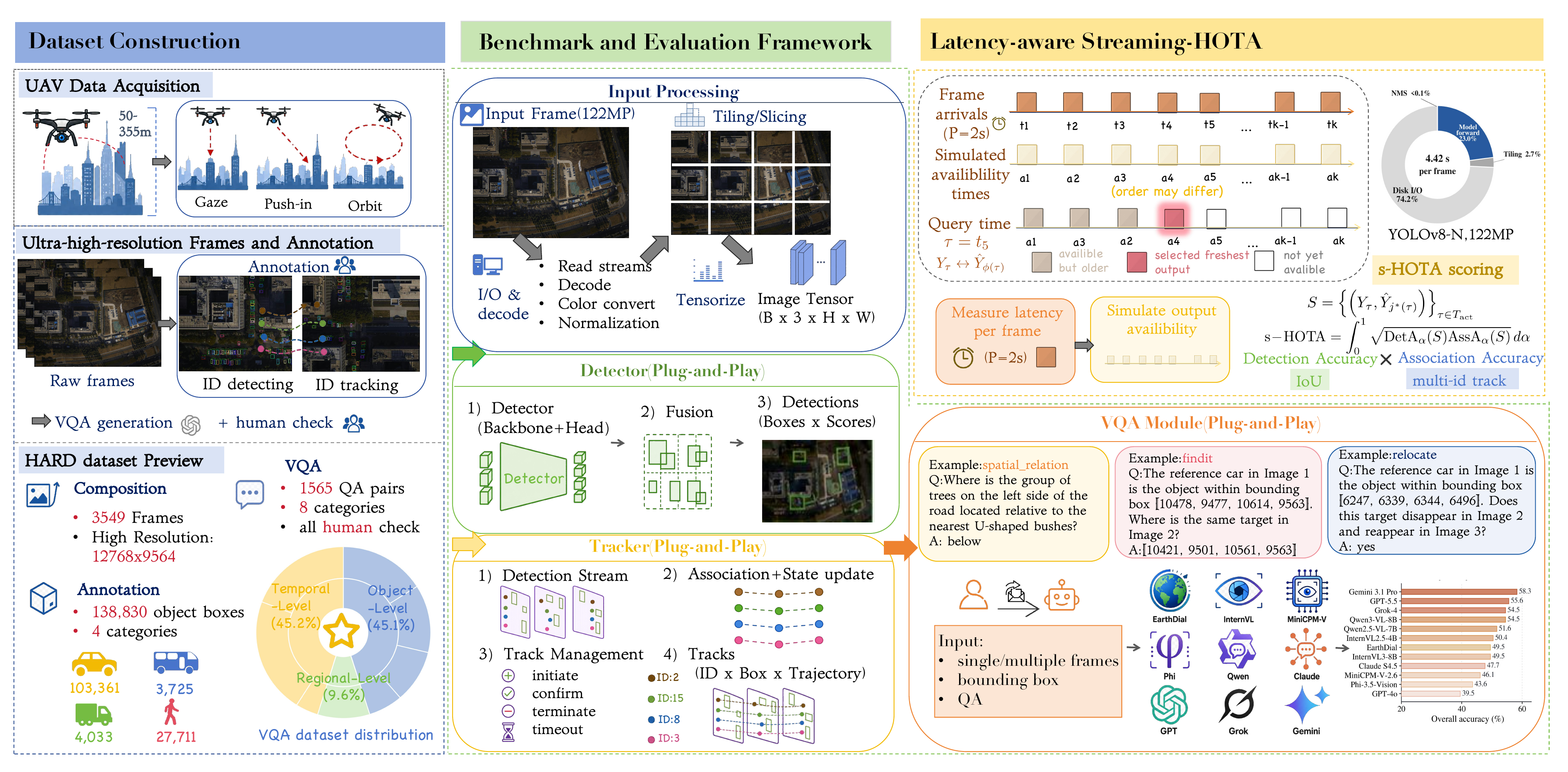}
  \caption{Overview of the HARD benchmark.
  Left: dataset construction, covering 122\,MP data collection over 21 sequences, 50--355\,m altitudes, and four flight modes, the three-level annotation hierarchy (detection, tracking, scene-level VQA). Middle: the benchmark and evaluation framework, in which detectors, trackers, and vision-language models are evaluated on the three-level tasks. Right: the definition and computation of the latency-aware streaming-HOTA metric.}
  \label{fig:main}
\end{figure*}

\paragraph{Object Detection and Multi-object Tracking Datasets.}
Datasets for aerial object detection and tracking have grown substantially over the past decade.
Figure~\ref{fig:dataset_comparison} places them against HARD in resolution, platform, temporal continuity and annotation coverage.
On the UAV platform, VisDrone~\cite{zhu2021detection} and UAVDT~\cite{du2018unmanned} provide rich object annotations and video streams covering diverse urban targets.
Their frame resolutions of about 3\,MP and 0.55\,MP, however, are far too small to expose the challenges of ultra-high-resolution perception.
On the satellite platform, VISO~\cite{viso2021} and SAT-MTB~\cite{satmtb2023} are likewise limited by small image sizes.
DOTA~\cite{ding2021dota} and FAIR1M~\cite{sun2022fair1m} contain some large images but offer no temporal data. In fact, 88\% of DOTA-v2.0 images are $1024{\times}1024$ patches and 45.6\% of FAIR1M images are only $800{\times}600$.
PANDA~\cite{panda2020} does offer gigapixel resolution and temporal sequences. 
But its cameras sit on the ground and look outward at a slanted angle, so the same type of target appears much larger up close than far away.
This differs from the near-vertical view that remote sensing works with.
Unlike these datasets~\cite{satmtb2023,ding2021dota,panda2020}, HARD unifies a UAV platform, 122\,MP resolution, temporal sequences, and multi-class tracking in a single corpus, and further adds scene-level VQA annotations.

\paragraph{Remote Sensing Visual Question Answering.}
In recent years, the visual question answering task has been extended to remote sensing.
RSVQA~\cite{rsvqa2020} pioneered auto-generated QA from OpenStreetMap annotations, and later work scaled this approach to large satellite corpora such as RSVQAxBEN~\cite{rsvqaxben} and RSIVQA~\cite{rsivqa}. 
The same approach has also been adapted to narrower settings including bi-temporal change reasoning (CDVQA~\cite{cdvqa2022}) and urban-planning relational reasoning (EarthVQA~\cite{earthvqa2024}).
Aerial and UAV-oriented benchmarks have followed, yet none targets wide-area dynamic traffic reasoning at ultra-high resolution.
HRVQA~\cite{li2024hrvqa} offers over a million QA pairs but on static $1024{\times}1024$ scene semantics.
Traffic-VQA~\cite{trafficvqa} addresses traffic cognition but relies on paired optical-thermal inputs and regulation-centric knowledge.
UAVReason~\cite{uavreason} evaluates broad UAV multimodal capability rather than urban situational awareness.
Existing remote-sensing VQA benchmarks ask questions about small static images and cannot support dynamic perception over wide areas. 
HARD covers wide areas at ultra-high resolution and provides cross-frame question-answer pairs.
It moves remote-sensing VQA from single small still images to cross-frame questions asked over 122\,MP wide-area frames. 

\paragraph{Evaluation Metrics for Detection and Tracking.}
Detection is measured by average precision~\cite{lin2014coco}, while tracking must also credit identity consistency.
MOTA~\cite{bernardin2008clear} and IDF1~\cite{ristani2016performance} are each dominated by a single aspect of tracking performance. 
HOTA~\cite{luiten2021hota} balances detection and association accuracy in one score, and we adopt it as our base metric.
All of these metrics are offline and do not model latency, scoring each prediction against its frame regardless of when it could be produced.
\citet{li2020streaming} showed that this assumption breaks for real-time systems and proposed streaming perception (sAP).
\citet{sela2022context} optimize a streaming variant of MOTA called S-MOTA when scheduling pipeline configurations for autonomous driving and \citet{peng2025towards} extend the streaming setting to 3D perception.
Different from all of them, our s-HOTA is an online, system-level metric for the end-to-end tracking pipeline.
Offline tracking evaluation rests on a zero-latency assumption that breaks down once one frame takes seconds to process.
s-HOTA corrects for this by scoring each frame against the latest prediction available at that time.

\section{The HARD Benchmark}
\label{sec:dataset}

HARD is designed not merely as a collection of images but as a research platform for wide-area aerial scene understanding.
It jointly provides the wide field of view, per-target resolution, and temporal continuity that WSTU requires (Figure~\ref{fig:dataset_comparison}).
This section describes how the data is collected, annotated and turned into QA pairs, and reports the statistics of the dataset.

\paragraph{Data Collection.}
To keep every ground target identifiable across a wide field of view, we capture the imagery with a Phase One iXM-GS120, an ultra-high-resolution camera.
A UAV carrying this camera captures every frame at a unified $12768{\times}9564$ (122\,MP) resolution over two geographically distinct sites.
The flights span altitudes from 50\,m to 355\,m in more than ten steps.
They also follow four flight modes, including linear-nadir, linear-oblique, stationary-gaze, and orbit-gaze (Figure~\ref{fig:main}, left).
These modes together produce background translation, rotation, parallax, and static-background conditions.
We detail the viewing geometry of each mode in the supplementary material.
The captured frames are then organized into 21 sequences, each a single continuous flight and thus one temporally ordered image sequence, with 169 frames per sequence on average.
This protocol yields a wide spread of object scales, viewing perspectives, and motion conditions within a single dataset.

\paragraph{Detection and Tracking Annotation.}
Annotation follows a two-tier protocol.
Firstly, five annotators in related fields holding master's degrees label every visible target with a tight box over its extent.
Each box follows a YOLO-normalized format extended with a per-instance identity field $(C_{\mathrm{id}}, I_{\mathrm{id}}, c_x, c_y, w, h)$.
The boxes span four categories, including pedestrian, car, bus, and truck.
Secondly, a more senior researcher audits each batch, correcting misclassifications, refining imprecise boxes, and returning substandard batches for re-annotation before acceptance.
Level~2 (tracking) uses the full annotation, whereas Level~1 (detection) reuses the same per-frame boxes without the cross-frame identity $I_{\mathrm{id}}$.
This process yields 138{,}830 annotations across 3{,}549 frames, with each annotator spending more than 100 hours on average and the full campaign costing over 600 hours of manual annotation and review.

\paragraph{VQA Generation.}
Beyond object detection and multi-object tracking, HARD further provides a VQA benchmark to evaluate whether vision--language models can progress from recognizing isolated targets to understanding wide-area dynamic scenes.
The benchmark is organized into three capability levels of increasing spatio-temporal complexity: single-frame perception, region-level understanding, and temporal target reasoning.
These levels respectively assess localized object recognition, evidence aggregation across an entire 122\,MP frame, and cross-frame reasoning over object identity and motion.
The benchmark construction combines annotation-derived supervision with assistance from GPT-5.5~\cite{gpt55}.
Bounding boxes and instance id trajectories are used to deterministically derive or ground questions and answers whenever possible, while the model assists in formulating natural-language questions and constructing plausible distractors.
All generated samples are subsequently reviewed by four experts in remote-sensing and UAV image interpretation, who verify visual answerability, option distinguishability, and annotation--answer consistency.
The final benchmark contains 1{,}565 manually verified question--answer pairs spanning eight question types.
Detailed task definitions, generation procedures, and type-wise statistics are provided in the supplementary material.

\paragraph{Dataset Statistics.}
HARD comprises 3{,}549 frames and 138{,}830 object annotations across the 21 sequences, an average of 39.1 objects per frame.
Of the 21 sequences, five are held out as the test set for all benchmarks, while the remaining sequences are used for training and validation.
The class distribution is car-dominant (74.4\%, 103{,}361 instances), followed by pedestrian (20.0\%, 27{,}711), bus (2.9\%, 4{,}033), and truck (2.7\%, 3{,}725).
This long-tailed distribution is consistent with the composition of everyday traffic scenes and poses a realistic class-imbalance challenge for detection and tracking.
A defining characteristic of HARD is its scale distribution, in which tiny and small targets together account for 98.7\% of all annotations (24.6\% tiny + 74.1\% small), each occupying between 0.001\% and 0.25\% of the frame.
Yet the 122\,MP resolution preserves enough pixels per target to keep individual instances distinguishable, a property lower-resolution UAV datasets cannot offer.

\noindent\textbf{Ethics and privacy.}
All flights were conducted under the applicable local UAV regulations. 
The acquisition geometry does not preserve resolvable facial or license-plate information, and the released annotations contain no biometric, demographic, or cross-scene identity labels. 
Further details are provided in the supplementary material.

\providecommand{\detrankfirst}[1]{%
  \cellcolor[RGB]{246,204,203}\textbf{#1}%
}
\providecommand{\detranksecond}[1]{%
  \cellcolor[RGB]{244,220,187}\textbf{#1}%
}
\providecommand{\detrankthird}[1]{%
  \cellcolor[RGB]{246,240,191}\textbf{#1}%
}

\begin{table*}[!t]
\centering
\footnotesize
\setlength{\tabcolsep}{1.5pt}
\renewcommand{\arraystretch}{1.08}

\begin{tabular}{@{}lcrrrcccccccccc@{}}
\toprule

\multirow{2}{*}{\textbf{Detector}}
&
\multirow{2}{*}{\textbf{Backbone}}
&
\multirow{2}{*}{\textbf{Params}}
&
\multirow{2}{*}{\textbf{Total}}
&
\multirow{2}{*}{\textbf{Infer}}
&
\multicolumn{2}{c}{\textbf{Pedestrian}}
&
\multicolumn{2}{c}{\textbf{Car}}
&
\multicolumn{2}{c}{\textbf{Bus}}
&
\multicolumn{2}{c}{\textbf{Truck}}
&
\multicolumn{2}{c}{\textbf{mAP}}
\\

\cmidrule(lr){6-7}
\cmidrule(lr){8-9}
\cmidrule(lr){10-11}
\cmidrule(lr){12-13}
\cmidrule(lr){14-15}

&
&
&
&
&
$\mathrm{AP}_{50}$
&
$\mathrm{AP}_{50:95}$
&
$\mathrm{AP}_{50}$
&
$\mathrm{AP}_{50:95}$
&
$\mathrm{AP}_{50}$
&
$\mathrm{AP}_{50:95}$
&
$\mathrm{AP}_{50}$
&
$\mathrm{AP}_{50:95}$
&
$\mathrm{AP}_{50}$
&
$\mathrm{AP}_{50:95}$
\\

\midrule

\multicolumn{15}{l}{\textit{Two-stage detectors}} \\

Faster R-CNN
& ResNet-50
& 41.8M
& 14.39
& 10.97
& 0.287
& 0.101
& \detranksecond{0.913}
& 0.406
& 0.222
& 0.110
& \detranksecond{0.780}
& \detrankfirst{0.466}
& \detrankthird{0.551}
& \detranksecond{0.271}
\\

Cascade R-CNN
& ResNet-50
& 69.2M
& 16.89
& 13.47
& 0.299
& 0.098
& \detrankthird{0.910}
& \detrankthird{0.423}
& \detranksecond{0.524}
& \detranksecond{0.260}
& \detrankfirst{0.866}
& \detranksecond{0.440}
& \detrankfirst{0.650}
& \detrankfirst{0.305}
\\

\midrule

\multicolumn{15}{l}{\textit{One-stage detectors}} \\

RetinaNet
& ResNet-50
& 37.7M
& 15.07
& 11.65
& 0.313
& 0.113
& \detrankfirst{0.915}
& \detranksecond{0.426}
& 0.267
& 0.146
& 0.144
& 0.072
& 0.410
& 0.189
\\

CenterNet
& ResNet-18
& 14.2M
& 8.85
& 5.43
& \detrankfirst{0.384}
& \detranksecond{0.135}
& 0.801
& 0.367
& 0.125
& 0.064
& 0.004
& 0.001
& 0.329
& 0.142
\\

YOLOv5s
& CSPDarknet-S
& 7.2M
& 5.41
& 1.99
& 0.211
& 0.061
& 0.809
& 0.378
& 0.376
& 0.186
& 0.417
& 0.139
& 0.453
& 0.191
\\

YOLOv5m
& CSPDarknet-M
& 21.2M
& 5.52
& 2.10
& 0.214
& 0.054
& 0.813
& 0.383
& 0.473
& 0.256
& 0.354
& 0.121
& 0.464
& 0.203
\\

YOLOv5l
& CSPDarknet-L
& 46.5M
& 6.16
& 2.74
& 0.326
& 0.091
& 0.824
& 0.386
& \detrankthird{0.519}
& \detranksecond{0.270}
& 0.141
& 0.060
& 0.453
& 0.202
\\

YOLOv8n
& CSPNet-N
& \detrankthird{3.2M}
& \detrankfirst{4.87}
& \detrankfirst{1.45}
& 0.223
& 0.066
& 0.827
& 0.390
& 0.422
& 0.214
& 0.313
& 0.105
& 0.446
& 0.194
\\

YOLOv8m
& CSPNet-M
& 25.9M
& 5.94
& 2.52
& 0.202
& 0.059
& 0.816
& 0.380
& 0.462
& 0.223
& 0.207
& 0.058
& 0.422
& 0.180
\\

YOLOv8l
& CSPNet-L
& 43.7M
& 6.50
& 3.08
& 0.222
& 0.050
& 0.805
& 0.380
& 0.389
& 0.210
& 0.660
& 0.220
& 0.519
& 0.215
\\

YOLO11n
& C3k2-N
& \detranksecond{2.6M}
& \detranksecond{4.96}
& \detranksecond{1.53}
& 0.133
& 0.027
& 0.826
& 0.386
& 0.356
& 0.187
& 0.313
& 0.105
& 0.407
& 0.176
\\

YOLO11m
& C3k2-M
& 20.1M
& 5.59
& 2.17
& 0.102
& 0.018
& 0.819
& 0.379
& 0.486
& \detrankfirst{0.284}
& 0.650
& 0.221
& 0.514
& 0.226
\\

YOLO11l
& C3k2-L
& 25.3M
& 5.94
& 2.52
& 0.071
& 0.014
& 0.760
& 0.349
& 0.451
& 0.226
& \detrankthird{0.729}
& 0.240
& 0.503
& 0.207
\\

YOLO26n
& CSPDarknet
& \detrankfirst{2.4M}
& \detrankthird{4.97}
& \detrankthird{1.55}
& 0.189
& 0.046
& 0.779
& 0.379
& 0.048
& 0.023
& 0.000
& 0.000
& 0.254
& 0.112
\\

YOLO26m
& CSPDarknet
& 20.4M
& 5.85
& 2.43
& 0.109
& 0.020
& 0.814
& 0.382
& 0.407
& 0.207
& 0.676
& 0.210
& 0.501
& 0.205
\\

YOLO26l
& CSPDarknet
& 24.8M
& 5.92
& 2.50
& 0.175
& 0.039
& 0.860
& 0.415
& 0.411
& 0.207
& 0.490
& 0.176
& 0.484
& 0.209
\\

\midrule

\multicolumn{15}{l}{\textit{Transformer-based detectors}} \\

DINO
& ResNet-50
& 47.0M
& 26.41
& 22.99
& \detrankthird{0.334}
& \detrankthird{0.124}
& 0.894
& \detrankfirst{0.432}
& 0.342
& 0.197
& 0.288
& 0.146
& 0.465
& 0.225
\\

RT-DETR
& ResNet-50
& 32.0M
& 6.83
& 3.40
& 0.263
& 0.074
& 0.836
& 0.364
& \detrankfirst{0.614}
& \detrankthird{0.267}
& 0.568
& 0.195
& \detranksecond{0.570}
& 0.225
\\

D-FINE-N
& HGNetv2
& 3.8M
& 6.39
& 2.97
& 0.247
& 0.094
& 0.835
& 0.406
& 0.056
& 0.021
& 0.154
& 0.080
& 0.323
& 0.150
\\

D-FINE-L
& HGNetv2
& 31.2M
& 7.22
& 3.80
& \detranksecond{0.365}
& \detrankfirst{0.145}
& 0.836
& \detrankthird{0.408}
& 0.308
& 0.162
& 0.425
& \detrankthird{0.293}
& 0.484
& \detrankthird{0.252}
\\

\bottomrule
\end{tabular}

\caption{
Detection performance of baseline detectors on HARD.
Both $\mathrm{AP}_{50}$ and $\mathrm{AP}_{50:95}$ are reported per category
and as mean AP (mAP) over all four categories.
Light-red, light-orange, and light-yellow cells indicate the
first-, second-, and third-ranked results in each numeric column.
For Params, Total, and Infer, lower is better; for all AP metrics,
higher is better.
``Total'' and ``Infer'' are in seconds. ``Infer'' denotes net model forward-pass time;
``Total'' denotes end-to-end per-frame latency.\
}
\label{tab:detection_baseline}
\end{table*}

\section{Evaluation Metrics}
\label{sec:metrics}

\textbf{Standard Metrics.}
\label{sec:stdmetrics}
We evaluate detection under the COCO protocol~\cite{lin2014coco} and report $\mathrm{AP}_{50}$ and $\mathrm{AP}_{50:95}$, both per-category and mean.
For tracking we adopt HOTA~\cite{luiten2021hota}, which scores a tracker as the geometric mean of a detection-accuracy component (DetA) and an association-accuracy component (AssA), integrated over localization thresholds.
We prefer HOTA to MOTA~\cite{bernardin2008clear} and IDF1~\cite{ristani2016performance} because each of the latter is dominated by a single aspect, whereas HOTA weighs detection and identity association evenly, and its two components can be inspected separately.
Scene-level VQA is scored by multiple-choice accuracy, reported per question type and overall.

\textbf{Streaming-HOTA.}
\label{sec:shota}
Offline tracking evaluation assumes zero processing latency, treating the output computed from a frame as if it were available at that same frame.
This assumption fails on ultra-high-resolution streams.
A ultra-high-resolution frame must pass through loading, tiling, per-tile detection, cross-tile fusion, and association, so the end-to-end latency
$\Delta t_t = t^{\mathrm{load}}_t + t^{\mathrm{tile}}_t + t^{\mathrm{det}}_t + t^{\mathrm{nms}}_t + t^{\mathrm{assoc}}_t$
reaches seconds per frame, and by the time a result is available the scene has moved on.
We therefore propose streaming-HOTA (s-HOTA). 
Given a per-frame latency profile and an availability policy, s-HOTA pairs each ground-truth instant with the most recent output available at that instant and evaluates the pairing with HOTA. 
The policy is an explicit input rather than a fixed assumption.

We define $P$ as the interval between consecutive frames, which is determined by the imaging time of the camera. For our camera $P = 2\,\mathrm{s}$.
The lateness of the output computed from frame $t$ is
\begin{equation}
  k_t \;=\; \max\bigl(1,\ \lceil \Delta t_t / P \rceil\bigr),
  \label{eq:lateness}
\end{equation}
so that this output becomes available at frame $t+k_t$.
Eq.~\eqref{eq:lateness} states our default policy, under which every frame starts processing on arrival.
It is realized by $K$ workers, each a full replica of the benchmarked machine, with $K \ge \lceil \max_t \Delta t_t / P \rceil$ so that no frame ever queues.
Each worker runs one frame in isolation, so the latencies measured on a single device transfer to this deployment without contention.
s-HOTA charges a pipeline for the staleness.
Evaluation instant $t$ coincides with the capture time of frame $t$, so an output can never serve the frame it is computed from.

At every evaluation instant $\tau$, let $\varphi(\tau)$ denote the most recent output whose availability index does not exceed $\tau$.
s-HOTA is standard HOTA computed over the latency-adjusted pairing
\begin{equation}
  \mathcal{S} \;=\; \bigl\{\bigl(y_\tau,\ \hat{y}_{\varphi(\tau)}\bigr)\bigr\}_{\tau \in \mathcal{T}_{\mathrm{act}}},
  \label{eq:pairing}
\end{equation}
\begin{equation}
  \text{s-HOTA}
  \;=\;
  \int_{0}^{1}
  \sqrt{\mathrm{DetA}_{\alpha}(\mathcal{S})\cdot
        \mathrm{AssA}_{\alpha}(\mathcal{S})}\;
  \mathrm{d}\alpha ,
  \label{eq:shota}
\end{equation}
where $\mathcal{T}_{\mathrm{act}}$ excludes the warm-up instants before the first output is available.
s-HOTA attains its upper limit when every output is ready within one frame interval, in which case it reduces to offline HOTA evaluated with a one-frame lag.
As latency grows beyond $P$, the score degrades in frame-quantized steps governed by Eq.~\eqref{eq:lateness}.
The complete protocol and full deployment parameters are given in the supplementary material.

\begin{figure*}[!t]
    \centering
    \begin{subfigure}[t]{0.238\textwidth}
        \centering
        \includegraphics[width=\linewidth]{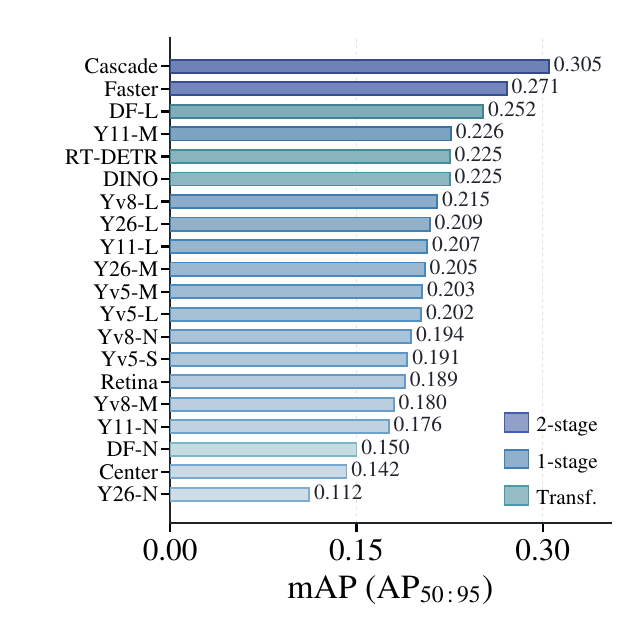}
        \caption{Detection ranking.}
        \label{fig:detection-ranking}
    \end{subfigure}
    \hfill
    \begin{subfigure}[t]{0.238\textwidth}
        \centering
        \includegraphics[width=\linewidth]{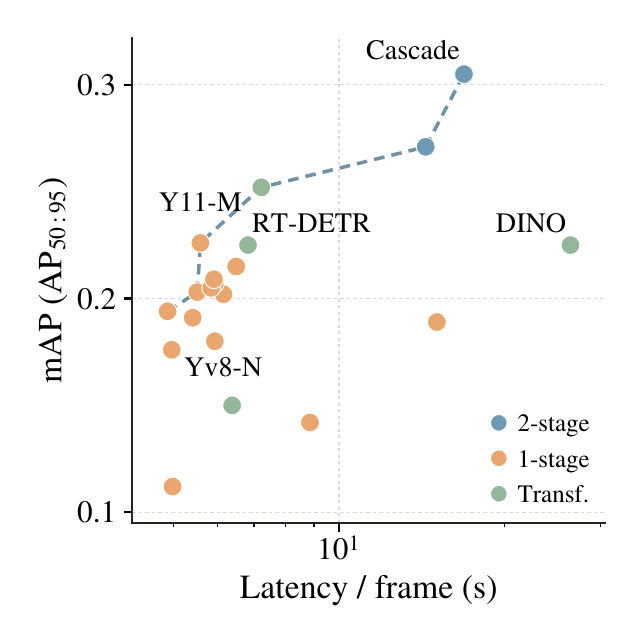}
        \caption{Accuracy--latency.}
        \label{fig:detection-latency}
    \end{subfigure}
    \hfill
    \begin{subfigure}[t]{0.238\textwidth}
        \centering
        \includegraphics[width=\linewidth]{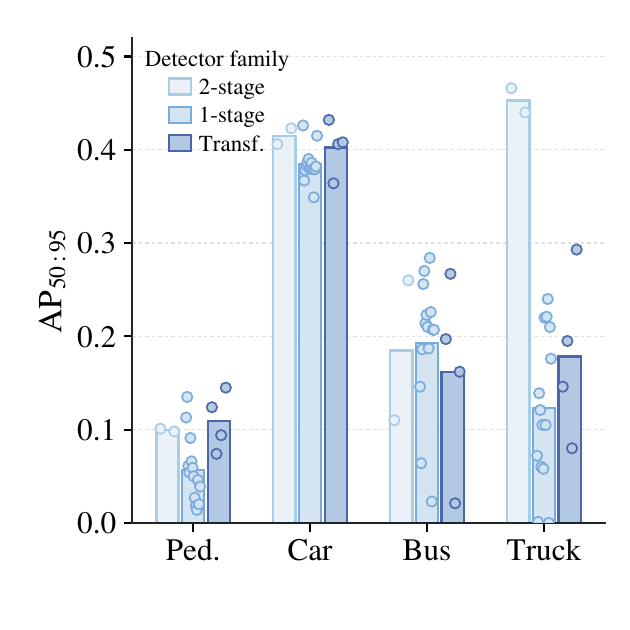}
        \caption{Class-wise landscape.}
        \label{fig:detection-classes}
    \end{subfigure}
    \hfill
    \begin{subfigure}[t]{0.238\textwidth}
        \centering
        \includegraphics[width=\linewidth]{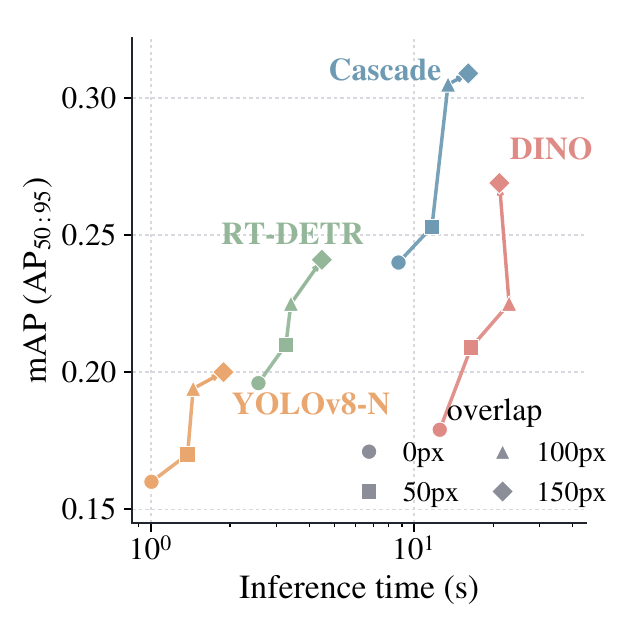}
        \caption{Overlap sensitivity.}
        \label{fig:detection-overlap}
    \end{subfigure}
    \caption{ Level-1 detection results:
        (a)~$\mathrm{AP}_{50:95}$ ranking of the twenty detectors;
        (b)~accuracy--latency trade-off with Pareto frontier;
        (c)~per-class $\mathrm{AP}_{50:95}$ by architectural family.
        For each object class, the three adjacent bars correspond, from left
        to right, to two-stage, one-stage, and transformer detectors.
        Bar heights give the mean AP of all detectors within each family,
        while the overlaid dots give the individual detector scores.
        The dots are slightly offset horizontally only to avoid visual
        overlap;
        (d)~effect of tile overlap on accuracy and inference time.}
    \label{fig:level1-detection}
\end{figure*}

\begin{table}[t]
\centering
\footnotesize
\setlength{\tabcolsep}{4.0pt}
\begin{tabular}{cccccc}
\toprule
\textbf{Overlap} & \textbf{Tiles} & \textbf{Obj./tile} & \textbf{Empty tiles} & \textbf{mAP} & \textbf{Infer} \\
\midrule
0\,px   & 300 & 0.130 & $\geq$87.0\% & 0.166 & 3.31\,s \\
50\,px  & 374 & 0.105 & $\geq$89.5\% & 0.178 & 4.32\,s \\
100\,px & 432 & 0.091 & $\geq$90.9\% & 0.204 & 5.25\,s \\
150\,px & 520 & 0.075 & $\geq$92.5\% & 0.213 & 5.90\,s \\
\bottomrule
\end{tabular}

\caption{Spatial sparsity of the HARD tiling pipeline under different overlaps, including tile density, empty-tile ratio, accuracy averaged over all twenty detectors, and inference time averaged over the eighteen detectors with directly measured per-overlap timings.}
\label{tab:sparsity}
\end{table}

\section{Experiments}
\label{sec:exp}

We evaluate HARD across three levels of increasing cognitive complexity (Figure~\ref{fig:main}, middle).
For each level, we first describe its experimental setup and then distill the results into its central findings. 

\subsection{Level 1: Object Detection}

\noindent\textbf{Setup.}
We benchmark twenty detectors spanning two-stage~\cite{ren2015faster,cai2018cascade}, one-stage~\cite{lin2017focal,zhou2019objects,jocher2020yolov5,jocher2023yolov8,jocher2024yolo11,ultralytics2026yolo26}, and Transformer-based~\cite{zhang2022dino,zhao2023rtdetr,peng2024dfine} families, all trained on $640{\times}640$ tiles with COCO-pretrained weights and evaluated under the COCO protocol with 100\,px inference overlap.
Table~\ref{tab:detection_baseline} reports the full results, Figure~\ref{fig:level1-detection} visualizes them.

\noindent\textbf{Finding 1: wide-area detection is shaped as much by the end-to-end tiling pipeline as by the detector, a joint consequence of extreme image size and target sparsity.}
Table~\ref{tab:sparsity} quantifies a property that aggregate mAP hides.
At the main configuration, a frame is split into 432 tiles yet contains only 39.1 objects on average, so most tiles are pure background.
Widening the tile overlap from 0 to 150\,px raises mean mAP from 0.166 to 0.213 by recovering targets fragmented at tile borders. 
The cost is that the tile count grows from 300 to 520 and the mean forward pass grows from 3.3\,s to 5.9\,s (Figure~\ref{fig:detection-overlap}).
The accuracy--latency landscape in Figure~\ref{fig:detection-latency} is thus a property of complete pipelines rather than of detectors alone.
The detector fixes the accuracy attainable on each tile, while the tiling and fusion stages decide how much of it survives at frame scale and at what latency.
Wide-area detection must therefore be designed as a joint system. Treating either component as fixed forfeits accuracy or latency that the other cannot recover.
Per-class results also diverge far more than aggregate mAP suggests (Figure~\ref{fig:detection-classes}).
Detectors agree closely on Car but spread from 0.000 to 0.466 $\mathrm{AP}_{50:95}$ on Truck.
 
\subsection{Level 2: Multi-Object Tracking}

\begin{figure}[!tb]
    \centering
    \begin{subfigure}[t]{0.49\columnwidth}
        \centering
        \includegraphics[width=\linewidth]{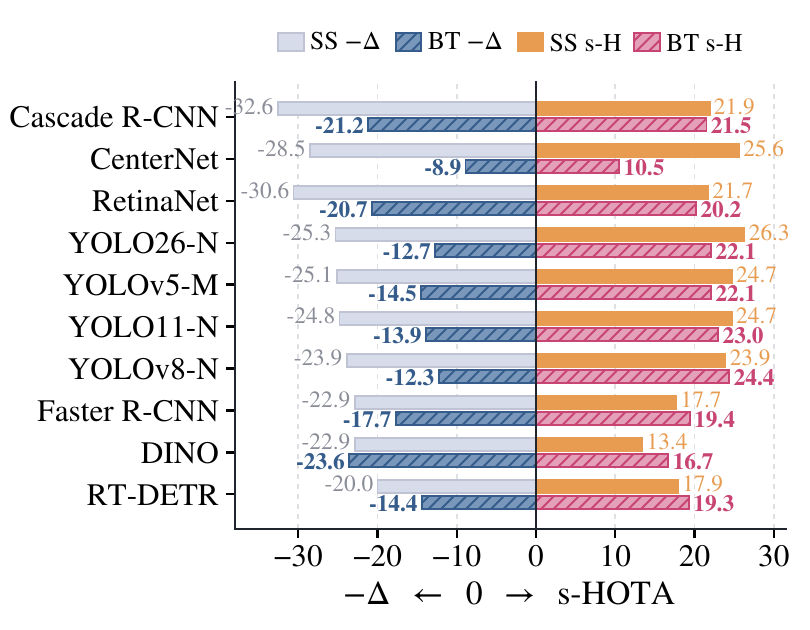}
        \caption{Score and latency penalty.}
        \label{fig:tracking-diverging}
    \end{subfigure}
    \hfill
    \begin{subfigure}[t]{0.49\columnwidth}
        \centering
        \includegraphics[width=\linewidth]{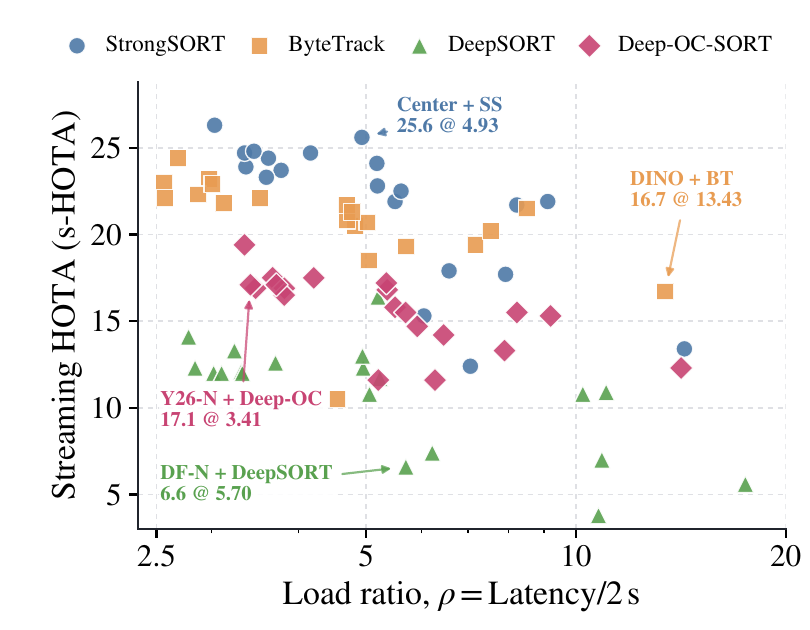}
        \caption{Load ratio vs.\ s-HOTA.}
        \label{fig:tracking-latency}
    \end{subfigure}
    \caption{Level-2 streaming-tracking results: (a)~s-HOTA and streaming penalty $-\Delta$ of StrongSORT vs.\ ByteTrack across representative detectors; (b)~s-HOTA against the load ratio $\rho=\Delta t/P$ over all 80 detector--tracker pipelines.}
    \label{fig:level2-tracking}
\end{figure}

\textbf{Setup.}
We pair every detector with four tracking methods, namely DeepSORT~\cite{wojke2017simple}, ByteTrack~\cite{zhang2022bytetrack}, Deep-OC-SORT~\cite{maggiolino2023deep}, and StrongSORT~\cite{du2023strongsort}, all under their official default hyperparameters.
StrongSORT therefore keeps its default sparse-optical-flow camera-motion compensation while the other three run without external compensation.
The supplementary material isolates the effect of this component.
This gives 80 pipelines, which we evaluate with s-HOTA at $P=2$\,s on the Car category (Figure~\ref{fig:level2-tracking}a). 
We choose this category because its dense and long trajectories make it the most statistically reliable one for streaming comparison. Table~\ref{tab:streaming_tracking} reports HOTA, s-HOTA, the streaming penalty $\Delta$ and end-to-end latency.

\textbf{Finding 2: streaming latency charges the association axis far more unevenly than the detection axis, so identity association is where tracking pipelines diverge under latency.}
Decomposing every score into HOTA's detection and association components shows that streaming charges both.
The two axes differ, however, in how much they separate the trackers.
Averaged over the twenty detector blocks, the fraction of offline DetA that survives streaming spans only 39--50\% across the four trackers.
The surviving fraction of AssA spans 53--88\%.
Association retention therefore varies about three times as widely as detection retention.
StrongSORT and Deep-OC-SORT, which maintain appearance feature banks, retain roughly half of their offline AssA (55\% and 53\%).
ByteTrack, which associates on motion and detection confidence alone, retains 65\%.
DeepSORT retains 88\%, but from a much lower offline base of 24--28 AssA against 53--66 for StrongSORT.
Its retention does not measure robustness of a comparable quantity.
Separating the contributions of appearance embeddings, camera-motion compensation and the motion model would require varying each in isolation.
On ultra-high-resolution sequences the practical lever is therefore sustaining identity across a stale gap rather than improving box quality.

\textbf{Finding 3: which tracking pipeline performs best under streaming is governed by the load ratio $\rho=\Delta t/P$.}
The net value of appearance cues shrinks as $\rho$ grows, so the best pipeline depends on the deployment budget rather than on the tracker alone.
An inter-frame-interval sweep isolates this effect. Tightening the frame budget from $P=2.5$ to $1$\,s with all predictions held fixed flips the s-HOTA winner from StrongSORT to ByteTrack on the high-latency two-stage and Transformer detectors. 
StrongSORT keeps its lead when $\rho$ stays small.
At $P=2$\,s, offline leads of up to 16.8 points compress to near-ties or reversals (Table~\ref{tab:streaming_tracking}). The best pipeline (StrongSORT on YOLO26-N) sits at low $\rho$ rather than at the offline optimum.
The association-insensitive S-MOTA cannot expose the differential at all.
The choice of tracking pipeline should therefore follow the deployment load ratio. Appearance-based association pays off only when latency room remains within the frame interval.

\providecommand{\trackdown}[1]{%
  \textcolor[RGB]{180,60,60}{$\downarrow$#1}%
}

\providecommand{\trackup}[1]{%
  \textcolor[RGB]{45,125,65}{$\uparrow$#1}%
}

\providecommand{\tracksame}[1]{%
  \textcolor[RGB]{150,150,150}{$\rightarrow$#1}%
}

\providecommand{\trackrankfirst}[1]{%
  \begingroup
  \fboxsep=1.2pt%
  \colorbox[RGB]{246,204,203}{\strut\textbf{#1}}%
  \endgroup
}

\providecommand{\trackranksecond}[1]{%
  \begingroup
  \fboxsep=1.2pt%
  \colorbox[RGB]{244,220,187}{\strut\textbf{#1}}%
  \endgroup
}

\providecommand{\trackrankthird}[1]{%
  \begingroup
  \fboxsep=1.2pt%
  \colorbox[RGB]{246,240,191}{\strut\textbf{#1}}%
  \endgroup
}

\begin{table*}[t]
\centering
\footnotesize
\setlength{\tabcolsep}{2.0pt}
\renewcommand{\arraystretch}{1.10}

\begin{tabular}{@{}l *{4}{ccc}@{}}
\toprule

\multirow{2}{*}{\textbf{Detector}}
&
\multicolumn{3}{c}{\textbf{StrongSORT}}
&
\multicolumn{3}{c}{\textbf{ByteTrack}}
&
\multicolumn{3}{c}{\textbf{DeepSORT}}
&
\multicolumn{3}{c}{\textbf{Deep-OC-SORT}}
\\

\cmidrule(lr){2-4}
\cmidrule(lr){5-7}
\cmidrule(lr){8-10}
\cmidrule(lr){11-13}

&
\textbf{HOTA}$\uparrow$
&
\textbf{s-HOTA}$\uparrow$
&
\textbf{Lat.}$\downarrow$
&
\textbf{HOTA}$\uparrow$
&
\textbf{s-HOTA}$\uparrow$
&
\textbf{Lat.}$\downarrow$
&
\textbf{HOTA}$\uparrow$
&
\textbf{s-HOTA}$\uparrow$
&
\textbf{Lat.}$\downarrow$
&
\textbf{HOTA}$\uparrow$
&
\textbf{s-HOTA}$\uparrow$
&
\textbf{Lat.}$\downarrow$
\\

\midrule

\multicolumn{13}{l}{\textit{Two-stage detectors}} \\

Faster R-CNN
& 40.6
& 17.7\,\trackdown{22.9}
& 15.85
& 37.1
& 19.4\,\trackdown{17.7}
& 14.36
& 11.7
& 7.0\,\trackdown{4.7}
& 21.79
& 29.5
& 13.3\,\trackdown{16.2}
& 15.80
\\

Cascade R-CNN
& \trackrankfirst{54.5}
& 21.9\,\trackdown{32.6}
& 18.22
& \trackrankfirst{42.7}
& 21.5\,\trackdown{21.2}
& 17.01
& \trackranksecond{20.3}
& 10.9\,\trackdown{9.4}
& 22.10
& \trackrankthird{36.8}
& 15.3\,\trackdown{21.4}
& 18.39
\\

\midrule

\multicolumn{13}{l}{\textit{One-stage detectors}} \\

RetinaNet
& \trackrankthird{52.3}
& 21.7\,\trackdown{30.6}
& 16.45
& \trackranksecond{40.9}
& 20.2\,\trackdown{20.7}
& 15.10
& 18.2
& 10.8\,\trackdown{7.5}
& 20.46
& \trackrankfirst{37.5}
& 15.5\,\trackdown{22.0}
& 16.46
\\

CenterNet
& \trackranksecond{54.0}
& \trackranksecond{25.6}\,\trackdown{28.5}
& 9.86
& 19.4
& 10.5\,\trackdown{8.9}
& 9.09
& \trackrankfirst{24.9}
& \trackrankfirst{16.4}\,\trackdown{8.5}
& 10.40
& 24.4
& 11.6\,\trackdown{12.8}
& 10.41
\\

YOLOv5-S
& 48.6
& 24.4\,\trackdown{24.2}
& 7.24
& 37.8
& \trackranksecond{23.2}\,\trackdown{14.7}
& 5.95
& 17.4
& 12.1\,\trackdown{5.4}
& 6.62
& 36.1
& \trackranksecond{17.5}\,\trackdown{18.6}
& 7.34
\\

YOLOv5-M
& 49.8
& 24.7\,\trackdown{25.1}
& 8.32
& 36.7
& 22.1\,\trackdown{14.5}
& 7.04
& 18.2
& 12.6\,\trackdown{5.6}
& 7.41
& 36.1
& \trackranksecond{17.5}\,\trackdown{18.5}
& 8.41
\\

YOLOv5-L
& 48.4
& 21.9\,\trackdown{26.4}
& 11.00
& 36.6
& 20.5\,\trackdown{16.2}
& 9.64
& 16.5
& 10.8\,\trackdown{5.6}
& 10.11
& 35.4
& 15.8\,\trackdown{19.6}
& 11.00
\\

YOLOv8-N
& 47.8
& 23.9\,\trackdown{23.9}
& \trackrankthird{6.72}
& 36.7
& \trackrankfirst{24.4}\,\trackdown{12.3}
& \trackrankthird{5.37}
& 17.0
& 12.0\,\trackdown{5.0}
& \trackrankthird{6.04}
& 35.8
& \trackrankfirst{19.4}\,\trackdown{16.4}
& \trackrankfirst{6.69}
\\

YOLOv8-M
& 48.3
& 23.7\,\trackdown{24.6}
& 7.55
& 36.6
& 21.8\,\trackdown{14.9}
& 6.25
& 17.4
& 12.0\,\trackdown{5.4}
& 6.64
& 35.4
& 16.9\,\trackdown{18.5}
& 7.63
\\

YOLOv8-L
& 48.8
& 22.5\,\trackdown{26.3}
& 11.22
& 36.9
& 20.7\,\trackdown{16.2}
& 10.04
& 17.5
& 11.7\,\trackdown{5.8}
& 10.46
& 35.3
& 15.5\,\trackdown{19.8}
& 11.39
\\

YOLO11-N
& 49.5
& 24.7\,\trackdown{24.8}
& \trackranksecond{6.69}
& 36.9
& \trackrankthird{23.0}\,\trackdown{13.9}
& \trackrankfirst{5.13}
& 17.6
& 12.3\,\trackdown{5.3}
& \trackranksecond{5.68}
& 36.3
& 16.9\,\trackdown{19.4}
& \trackrankthird{6.94}
\\

YOLO11-M
& 48.4
& 23.3\,\trackdown{25.2}
& 7.19
& 36.8
& 22.3\,\trackdown{14.5}
& 5.74
& 17.5
& 12.0\,\trackdown{5.5}
& 6.20
& 35.4
& 16.5\,\trackdown{18.9}
& 7.63
\\

YOLO11-L
& 48.2
& 22.8\,\trackdown{25.4}
& 10.38
& 36.9
& 20.8\,\trackdown{16.0}
& 9.39
& 18.2
& 12.3\,\trackdown{5.9}
& 9.90
& 36.1
& 16.8\,\trackdown{19.3}
& 10.72
\\

YOLO26-N
& 51.6
& \trackrankfirst{26.3}\,\trackdown{25.3}
& \trackrankfirst{6.06}
& 34.8
& 22.1\,\trackdown{12.7}
& \trackranksecond{5.14}
& \trackranksecond{20.3}
& \trackranksecond{14.1}\,\trackdown{6.2}
& \trackrankfirst{5.56}
& 36.0
& 17.1\,\trackdown{18.8}
& \trackranksecond{6.82}
\\

YOLO26-M
& 50.9
& \trackrankthird{24.8}\,\trackdown{26.0}
& 6.90
& 37.9
& 22.9\,\trackdown{15.0}
& 6.02
& \trackrankthird{19.7}
& \trackrankthird{13.3}\,\trackdown{6.3}
& 6.47
& 36.5
& 17.1\,\trackdown{19.4}
& 7.43
\\

YOLO26-L
& 51.7
& 24.1\,\trackdown{27.5}
& 10.36
& 38.4
& 21.7\,\trackdown{16.7}
& 9.38
& 19.5
& 13.0\,\trackdown{6.4}
& 9.88
& \trackranksecond{37.4}
& \trackrankthird{17.2}\,\trackdown{20.1}
& 10.69
\\

\midrule

\multicolumn{13}{l}{\textit{Transformer-based detectors}} \\

DINO
& 36.3
& 13.4\,\trackdown{22.9}
& 28.62
& \trackrankthird{40.3}
& 16.7\,\trackdown{23.6}
& 26.85
& 11.5
& 5.6\,\trackdown{5.9}
& 35.01
& 34.1
& 12.3\,\trackdown{21.8}
& 28.32
\\

RT-DETR
& 37.9
& 17.9\,\trackdown{20.0}
& 13.15
& 33.7
& 19.3\,\trackdown{14.4}
& 11.41
& 10.9
& 7.4\,\trackdown{3.5}
& 12.44
& 30.3
& 14.2\,\trackdown{16.1}
& 12.92
\\

D-FINE-N
& 32.6
& 15.3\,\trackdown{17.3}
& 12.10
& 37.6
& 21.3\,\trackdown{16.3}
& 9.55
& 9.7
& 6.6\,\trackdown{3.1}
& 11.40
& 31.1
& 14.7\,\trackdown{16.5}
& 11.84
\\

D-FINE-L
& 26.7
& 12.4\,\trackdown{14.3}
& 14.11
& 33.0
& 18.5\,\trackdown{14.5}
& 10.09
& 6.1
& 3.8\,\trackdown{2.2}
& 21.55
& 25.8
& 11.6\,\trackdown{14.1}
& 12.55
\\

\bottomrule
\end{tabular}

\caption{
\emph{Streaming-HOTA} on HARD (Car, $P=2$\,s), aggregated over all
sequences using TrackEval \textsc{combined}.
For each detector--tracker pipeline, we report offline HOTA,
streaming s-HOTA$\uparrow$, and end-to-end latency
Lat.(s)$\downarrow$.
The red downward value beside each s-HOTA score denotes its degradation relative to the corresponding offline HOTA.
Light-red, light-orange, and light-yellow boxes respectively indicate
the first-, second-, and third-ranked results in each metric column.
}
\label{tab:streaming_tracking}
\end{table*}

\subsection{Level 3: Scene-Level VQA}

\begin{figure}[!tb]
    \centering
    \begin{subfigure}[t]{0.49\columnwidth}
        \centering
        \includegraphics[width=\linewidth]{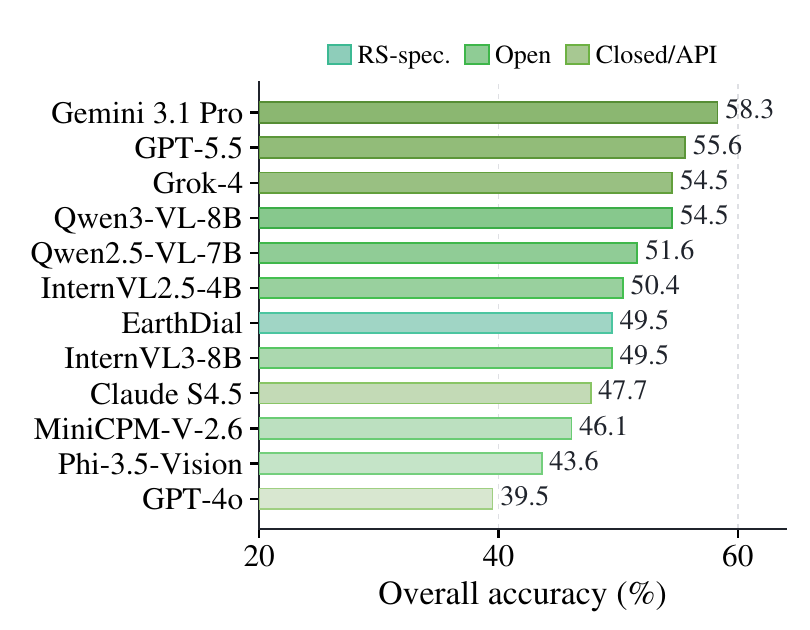}
        \caption{Overall ranking.}
        \label{fig:vqa-ranking}
    \end{subfigure}
    \hfill
    \begin{subfigure}[t]{0.49\columnwidth}
        \centering
        \includegraphics[width=\linewidth]{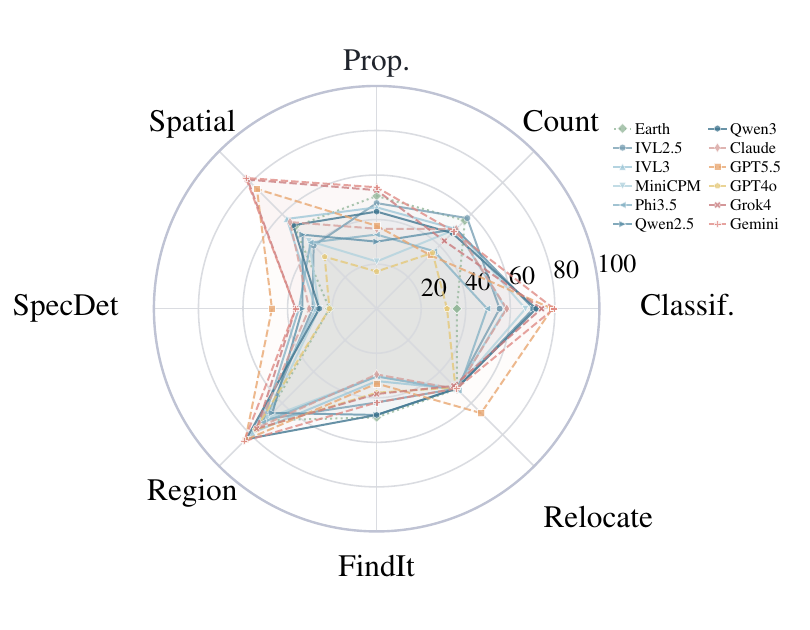}
        \caption{Per-task profiles.}
        \label{fig:vqa-profiles}
    \end{subfigure}
    \caption{Level-3 scene-level VQA results: (a)~overall accuracy ranking of the twelve vision-language models; (b)~per-task profiles across the eight question types.}
    \label{fig:level3-vqa}
\end{figure}

\noindent\textbf{Setup.}
We evaluate twelve vision-language models zero-shot on the 1{,}565 Level-3 questions, comprising one remote-sensing-specialized model~\cite{earthdial}, six open-source general-purpose models~\cite{internvl25,internvl3,minicpmv,phi35vision,qwen25vl,qwen3vl}, and five closed-source API models~\cite{gpt4o,gpt55,claude45,grok4,gemini31}.
Per-type accuracy is visualized in Figure~\ref{fig:level3-vqa}.

\noindent\textbf{Finding 4: VLM failures separate into two axes, resolution-limited perception and near-chance identity binding, and progress on one does not transfer to the other.}
Averaging the per-type accuracies into a single-frame axis (classification, counting, property, spatial relation) and a temporal axis (special detection, FindIt, Relocate) reveals two very different failure modes.
On the single-frame axis, models spread over a 37-point range (29.2\% for GPT-4o to 66.5\% for Gemini 3.1 Pro). This axis rewards general perception ability, and stronger models are clearly better despite the heavy downsampling that a 122\,MP input undergoes.
On the temporal axis, the same twelve models collapse into a 13-point band (35.6--49.0\%) hovering near the per-type chance levels, with eleven of twelve at 49--52\% on Relocate.
Only GPT-5.5 rises meaningfully above chance, reaching 66.3\% on Relocate and 47.0\% on special detection.
We report these as observations consistent with, rather than proof of, the two mechanisms.
The implication is that scaling general capability mainly moves the first axis, while closing the second likely requires explicit object memory or track-conditioned inputs, which is where Level-1 and Level-2 supervision in HARD becomes directly useful.

\section{Conclusion}

We introduced HARD, an ultra-high-resolution temporal UAV benchmark.
Its three-level annotations span object perception to scene cognition.
We also proposed s-HOTA to evaluate tracking under the second-scale latency such imagery induces.
Extensive baseline experiments surface two open problems that we believe warrant deeper study.
The first is how to lower the end-to-end latency of ultra-large images.
The second is how to give models persistent cross-frame identity.
We frame this line of work as Wide-area Spatio-temporal Scene Understanding.
We hope HARD can anchor future research toward real UAV deployment.


\bibliography{references}

\end{document}